\documentclass[10pt,twocolumn,letterpaper]{article}

\usepackage{ijcb}
\usepackage{times}
\usepackage{epsfig}
\usepackage{graphicx}
\usepackage{amsmath}
\usepackage{amssymb}

\usepackage{multirow}
\usepackage{algpseudocode}
\usepackage[ruled, lined, linesnumbered, commentsnumbered, longend]{algorithm2e}
\usepackage{subfigure}
\usepackage{diagbox}
\usepackage{arydshln}

\usepackage[pagebackref=true,breaklinks=true,letterpaper=true,colorlinks,bookmarks=false]{hyperref}

\ijcbfinalcopy 

\ifijcbfinal\fi
\begin{document}

\title{Unsupervised Face Morphing Attack Detection via Self-paced Anomaly Detection}

\author{Meiling Fang$^{1,2}$, Fadi Boutros$^{1}$, Naser Damer$^{1,2}$ \\
$^{1}$Fraunhofer Institute for Computer Graphics Research IGD,
Darmstadt, Germany\\
$^{2}$Department of Computer Science, TU Darmstadt,
Darmstadt, Germany\\
Email: {meiling.fang@igd.fraunhofer.de}
\vspace{-5mm}
}
\maketitle
\thispagestyle{empty}

\begin{abstract}
The supervised-learning-based morphing attack detection (MAD) solutions achieve outstanding success in dealing with attacks from known morphing techniques and known data sources. However, given variations in the morphing attacks, the performance of supervised MAD solutions drops significantly due to the insufficient diversity and quantity of the existing MAD datasets. To address this concern, we propose a completely unsupervised MAD solution via self-paced anomaly detection (SPL-MAD) by leveraging the existing large-scale face recognition (FR) datasets and the unsupervised nature of convolutional autoencoders.
Using general FR datasets that might contain unintentionally and unlabeled manipulated samples to train an autoencoder can lead to a diverse reconstruction behavior of attack and bona fide samples.
We analyze this behavior empirically to provide a solid theoretical ground for designing our unsupervised MAD solution.
This also results in proposing to integrate our adapted modified self-paced learning paradigm to enhance the reconstruction error separability between the bona fide and attack samples in a completely unsupervised manner.
Our experimental results on a diverse set of MAD evaluation datasets show that the proposed unsupervised SPL-MAD solution outperforms the overall performance of a wide range of supervised MAD solutions and provides higher generalizability on unknown attacks.
Training codes and pre-trained models are publicly released \footnote{\url{https://github.com/meilfang/SPL-MAD}}.


\end{abstract}

\vspace{-5mm}
\section{Introduction} 
Face recognition technique has witnessed remarkable progress in recent years. A variety of face recognition methods \cite{boutros2021elasticface,DBLP:conf/cvpr/DengGXZ19} are proposed in the literature and applied in practical applications with very high accuracy.
However, these methods are vulnerable to several attack \cite{DBLP:conf/bmvc/DamerD16,DBLP:conf/btas/DamerWBBT0K18,DBLP:journals/pr/FangDKK22,DBLP:journals/cviu/MassoliCAF21}, one of which is the face morphing attack. A morphing attack is a face image which is purposefully manipulated to be matched with the probe images of more than one identity. As a result, morphing attack detection (MAD) solutions is of crucial importance to building reliable face recognition systems. Conventional single-image and differential MAD solutions \cite{DBLP:conf/isvc/DamerSFBKK21,DBLP:books/sp/16/FerraraFM16,Ibsen2021,DBLP:conf/btas/RaghavendraRB16,DBLP:conf/cvpr/RaghavendraRVB17a,DBLP:conf/cvip/RamachandraVRB18} require two classes of data, bona fide (i.e., not attack) and morphing attack samples for training a supervised MAD model. However, this restricts the MAD performance with the size and diversity of the training data. Most of the existing MAD datasets \cite{DBLP:conf/btas/DamerS0K18,DBLP:conf/isvc/DamerSFBKK21, Sarkar2020} are limited in diversity and quantity caused by such as the labor-intensive pair selection, morphing process, and the limited of bona fide source data that is of suitable properties (e.g., ICAO compliant \cite{ICAO}) and shareable in a privacy-aware frame. Moreover, due to ethical and legal issues, only a few MAD datasets \cite{DBLP:conf/btas/DamerS0K18,DBLP:conf/isvc/DamerSFBKK21, Sarkar2020} are publicly available for the development of MAD solutions. Following the lack of diversity of MAD datasets and the possibility of facing attacks created by unknown methods, the supervised MAD solutions \cite{DBLP:conf/isvc/DamerSFBKK21,DBLP:books/sp/16/FerraraFM16,DBLP:conf/btas/RaghavendraRB16,DBLP:conf/cvpr/RaghavendraRVB17a,DBLP:conf/cvip/RamachandraVRB18} commonly results in poor performance generalization on unknown morphing attacks or data sources. 
To the best of our knowledge, only one single-image based \cite{DBLP:conf/btas/DamerGZKK19} and one differential based MAD method \cite{Ibsen2021} were proposed to detect morphing attacks as anomalies by training a one-class classifier. Despite the obtained improved performance on the unknown attack in comparison to supervised MAD approaches, training the one-class classifier still relies on the prior knowledge that all training samples are known to be bona fide, and thus it is not a completely unsupervised approach.

To target the lack of large-scale, labelled, and diverse MAD datasets, along with the low generalizability of supervised MADs on unknown attacks,  we leverage the existing large-scale face recognition datasets to train our proposed unsupervised learning-based model. 
Most publicly available face recognition datasets were collected from the web and might consist of unintentionally and unlabeled manipulated samples. To alleviate this issue, we model our design as self-paced learning (SPL) paradigm. Self-paced learning paradigm is inspired by the cognitive learning order in human curricula, where samples are involved in the training phase from easy to hard ones \cite{DBLP:conf/icml/BengioLCW09}. In this case, training data is evaluated and selected automatically based on the training loss without any prior knowledge of humans. Recently, researchers have investigated the potential of SPL paradigm \cite{DBLP:conf/aaai/FanHLH17,DBLP:conf/nips/KumarPK10} and demonstrated that is significant a strong performance gain \cite{DBLP:conf/eccv/XiangDH20, DBLP:journals/ijcv/ZhangHZM19}. Following the properties of the SPL paradigm as an effective learning strategy to suppress the side effects of noise samples or outliers by adjusting the weight of samples. Given the understanding of the limitations of supervised MAD solutions, we propose an SPL paradigm and incorporate it and incorporate it in our unsupervised MAD learning. 

This work makes the following main contributions: 1) We first study the behavior of unsupervised anomaly detection through reconstruction error analyses on MAD data to ensure that our unsupervised MAD solution is developed on the bases of solid empirical analyses. Our study reveals that morphing attacks are more straightforward to reconstruct than bona fide samples when the reconstruction is learned on general face data; 2) We leverage our above-stated observation to present a novel unsupervised MAD solution via an adapted self-paced anomaly detection, namely SPL-MAD. The adapted SPL paradigm proved helpful in neglecting the suspicious unlabeled data in training and thus enhancing the reconstruction gap between bona fide and attack samples, leading to improving the generalizability of the MAD model. 3) The experimental results demonstrate that our SPL-MAD solution not only reaches the performance of supervised MAD solutions but also outperforms well-established supervised MAD methods and presents a more generalizable performance over a diverse set of unknown attacks included in eight MAD datasets.


\section{Related work} 
\label{sec:related_work}

A number of studies \cite{DBLP:books/sp/16/FerraraFM16,DBLP:conf/iwbf/ScherhagRRGRB17} pointed out that face recognition systems are vulnerable to morphing attacks. To target this problem, several MAD solutions \cite{DBLP:conf/isvc/DamerSFBKK21,DBLP:conf/cvip/RamachandraVRB18,DBLP:conf/wacv/SoleymaniDTDN21} were proposed. MAD solutions can be categorized into two groups based on the application scenario requirements: single-image MAD and differential-MAD, where the latter requires an investigated image and an additional live capture of the individual \cite{DBLP:conf/dagm/DamerBWBTBK18}, which limits its applicability in many scenarios. In our case, we focus on the single-image-based MAD scenario, where only the investigated image is analyzed. 
Most single-image MAD solutions \cite{DBLP:conf/isvc/DamerSFBKK21,DBLP:books/sp/16/FerraraFM16,DBLP:conf/btas/RaghavendraRB16,DBLP:conf/cvpr/RaghavendraRVB17a,DBLP:conf/cvip/RamachandraVRB18} are based on supervised learning that relies on data annotations. For example, Raghavendra \etal \cite{DBLP:conf/cvip/RamachandraVRB18} proposed a handcrafted-feature-based solution where textural features were extracted across scale-space and were classified using collaborative representation. Damer \etal \cite{DBLP:conf/isvc/DamerSFBKK21} proposed a pixel-wise MAD (PW-MAD) solution where a network is trained to classify each pixel of the image into an attack or not, rather than only one binary label for the whole image. 
These supervised-learning-based solutions achieved good MAD performance typically on attacks with properties known during training. 
Variations in the attacks strongly effect the MAD performances, such variations can be related to the face morphing approach \cite{DBLP:conf/btas/DamerBSKK19,DBLP:conf/btas/DamerGZKK19,ReGenM,DBLP:conf/iciap/DebiasiDSRSBKU19,MIPGAN}, the pairing protocols of morphed images \cite{DBLP:conf/icb/DamerSZWTKK19,DBLP:conf/fusion/DamerZWSKK19}, image compression \cite{DBLP:conf/visapp/MakrushinND17}, the source of bona fide images \cite{DBLP:conf/iwbf/ScherhagRB18}, and re-digitization of the images \cite{MADVgg,DBLP:conf/btas/RaghavendraRB16,DBLP:conf/cvpr/RaghavendraRVB17a}, among other variations. 
Additionally, to the relative low generalizability of supervised MAD, their optimal training requires large-scale labelled databases with variation in the attacks, which is very challenging given the data-creating efforts and the legal limitations on using, sharing, and re-using biometric-based personal data \cite{gdpr}.

Besides supervised MAD approaches, a single work \cite{DBLP:conf/btas/DamerGZKK19} proposed to detect morphing attacks from single images by leveraging the anomaly detection technique, however, with very limited success. Damer \etal \cite{DBLP:conf/btas/DamerGZKK19} studied detecting attacks as anomalies via one-class classifiers. However, the performance of the one-class model on unknown attacks was still low, considering roughly 50\% detection error rates \cite{DBLP:conf/btas/DamerGZKK19}. Moreover, training the one-class classifier in fact relies on the prior knowledge that all training samples are assumed to be bona fide data even if they included contamination, and thus it is not a completely unsupervised approach. Also, using a one-class classifier, but for the out of our scope differential morphing attack detection that requires a bona fide (live) image in its operation, a recent work \cite{Ibsen2021} proposed a solution that also required bona fide labelled images for the training.

In addition to the lack of large-scale, labelled, and diverse datasets, leading to poor generalizability of the MAD model, most of the existing morph attack samples in MAD datasets \cite{DBLP:conf/btas/DamerS0K18, DBLP:conf/isvc/DamerSFBKK21, Sarkar2020} are created based on a small-scale bona fide samples. This is due to the insufficient identities in suitable (ICAO compliant \cite{ICAO}) and publicly available datasets. Compared to the face recognition datasets \cite{casiawebface}, and even face presentation attack (spoofing) databases \cite{DBLP:conf/eccv/ZhangYLYYSL20}, existing MAD datasets are hundred times smaller. Besides, only a few datasets (detailed information of eight MAD datasets can be found in Section \ref{ssec:datasets}) are available for MAD development research. To enhance the generalizability on unknown morphing attacks and avoid the need for diverse morphing development databases, we leverage the publicly available face recognition datasets and present an unsupervised face MAD solution, the SPL-MAD.



\section{Methodology} 
\label{sec:method}

\begin{figure*}[thbp]
    \centering
    \subfigure[]{\includegraphics[width=0.24\textwidth]{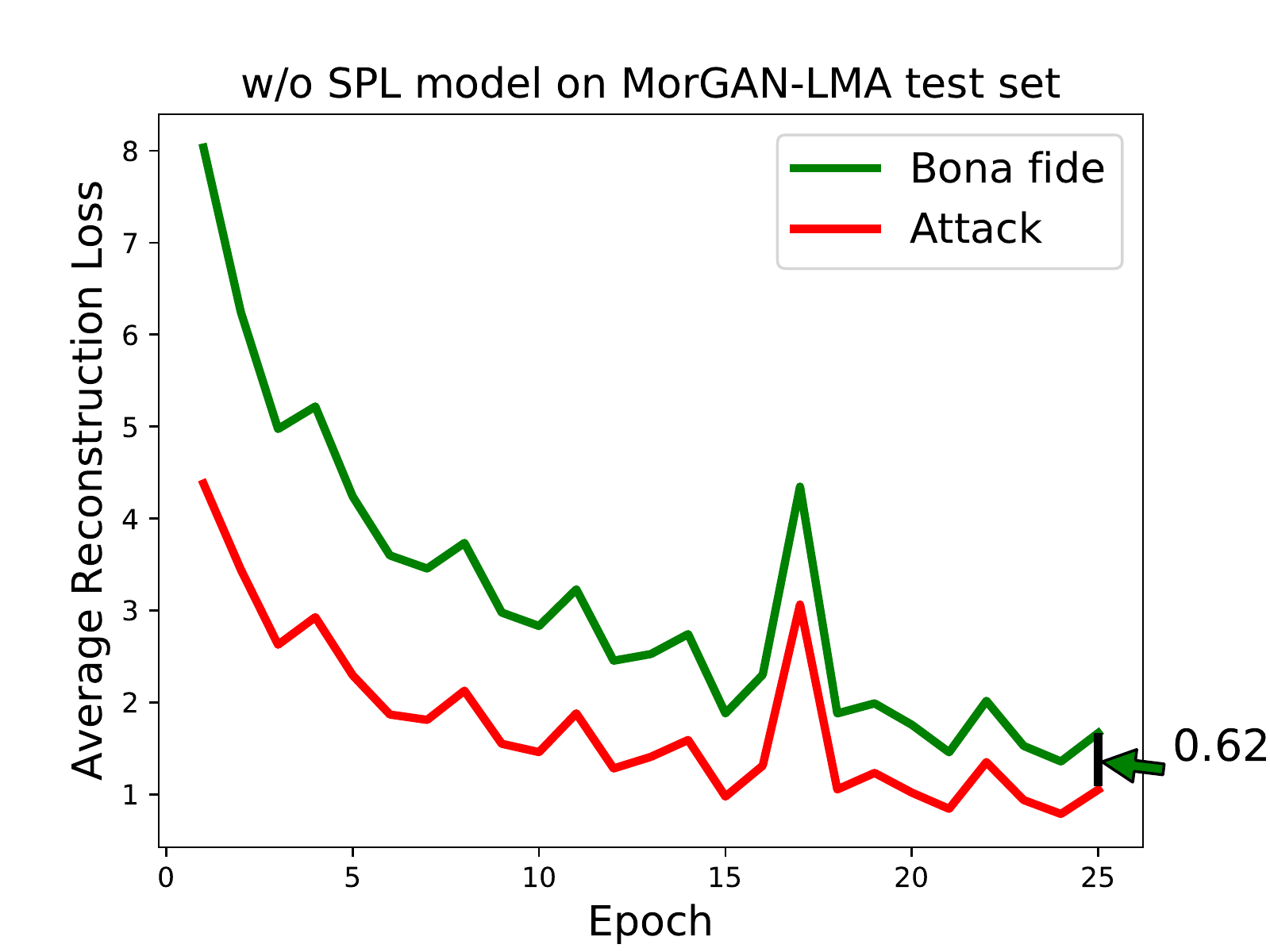}} 
    \subfigure[]{\includegraphics[width=0.24\textwidth]{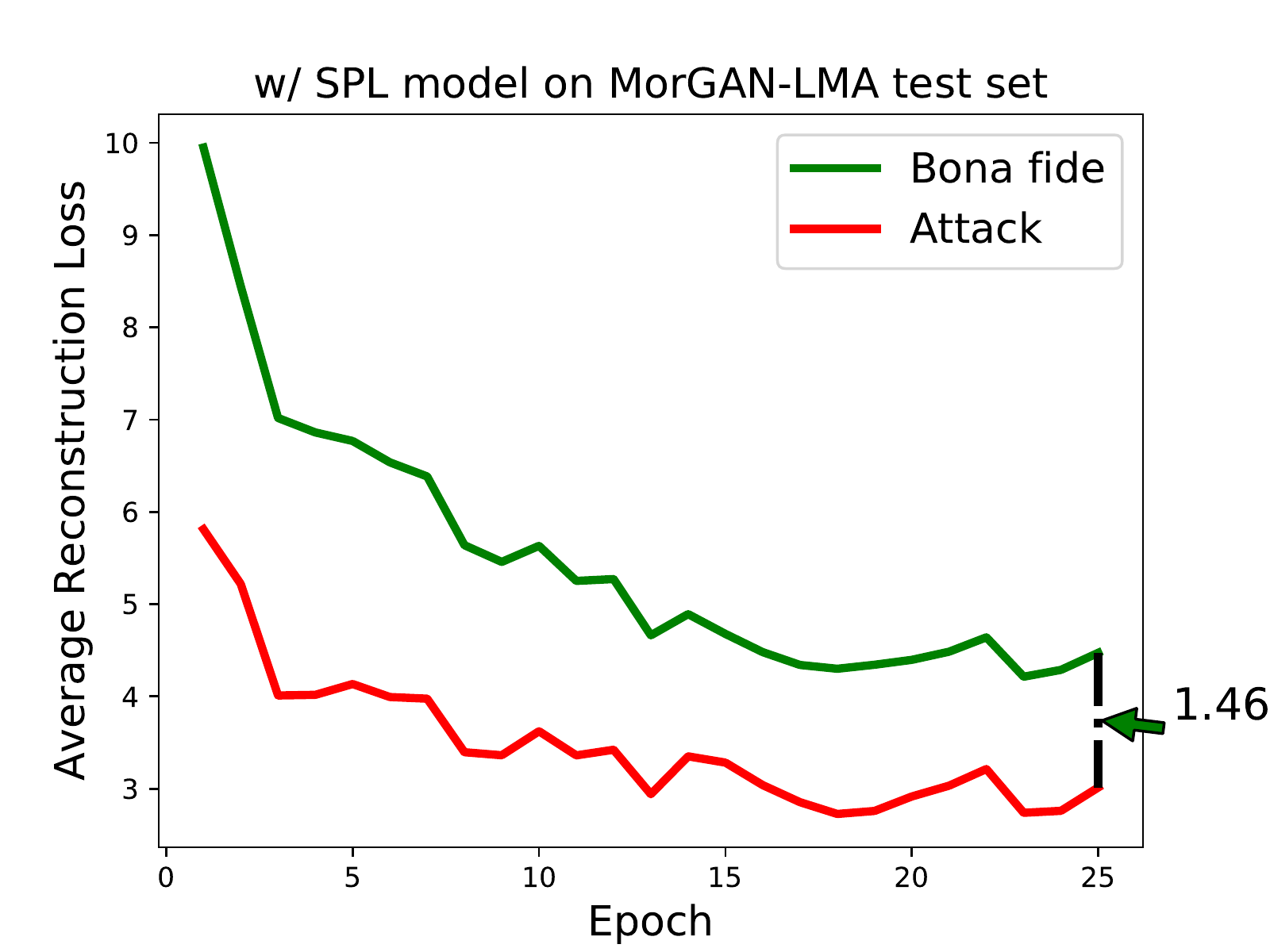}} 
    \subfigure[]{\includegraphics[width=0.24\textwidth]{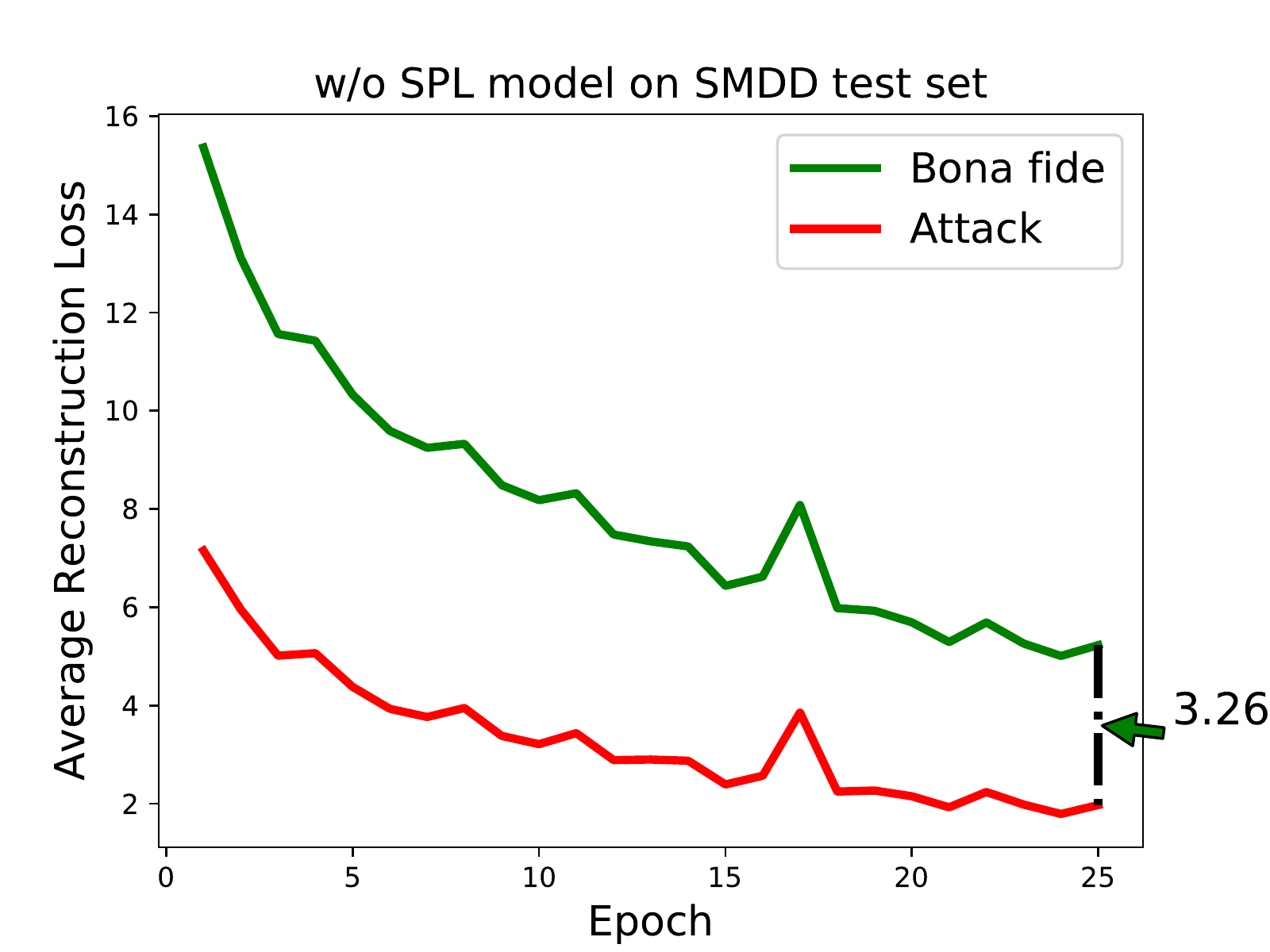}}
    \subfigure[]{\includegraphics[width=0.24\textwidth]{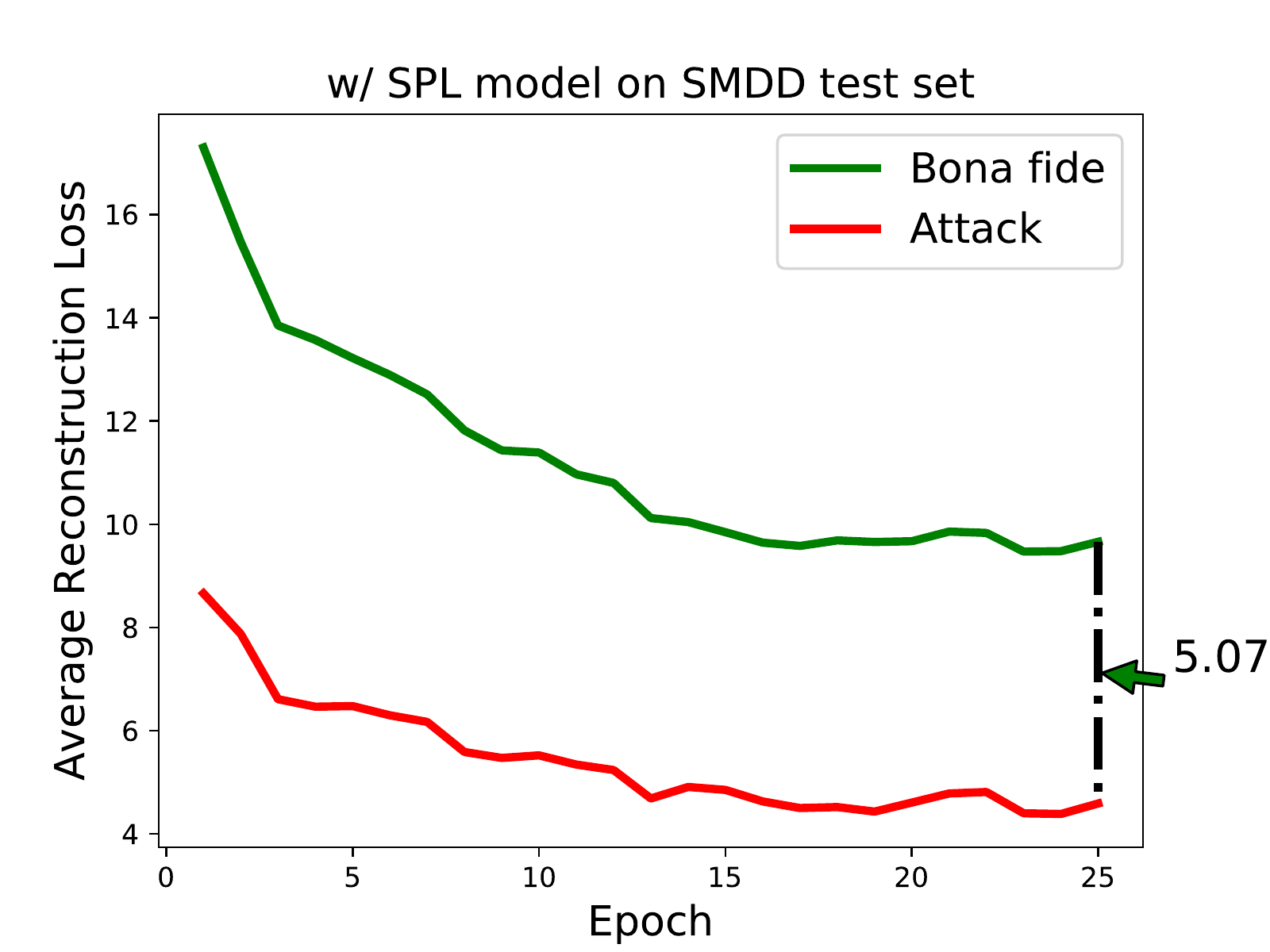}}
    \caption{The curves of reconstruction error on two MAD test set (SMDD \cite{damer2022privacy} and MorGAN-LMA \cite{DBLP:conf/btas/DamerS0K18}, details in Section \ref{ssec:datasets}) by models trained without SPL and with SPL paradigm. The x-axis refers to the training epoch, and the y-axis is the average reconstruction error of all data. The green curve denotes the error of bona fide data, and the red curve presents the error of attack data. It can be seen that attacks are easier to reconstruct than bona fide samples, thus resulting in lower reconstruction error, which we leverage in our proposed MAD solution. The SPL paradigm also leads to a higher gap in the reconstruction error between bona fide and attacks.}
    \vspace{-3mm}
    \label{fig:loss_curves}
\end{figure*}

\subsection{Preliminaries}
\label{ssec:Preliminaries}
\textbf{Convolutional autoencoder (CAE):}
Autoencoder (AE) is a branch of unsupervised learning techniques and has been widely used for anomaly detection \cite{soleymani2021mutual,DBLP:conf/iclr/ZongSMCLCC18}. AE consists of an encoder representing the input in a latent domain and a decoder reconstructing the data from this latent feature. Convolutional autoencoder (CAE) is designed by stacking convolutional layers, where the encoder (denote as $f_{e}(., \mathbf{w}_e)$) is combined by several convolutional layers and the decoder (denote as $f_{d}(., \mathbf{w}_d)$) is combined by transposed convolutional layers. Given $i$-th input $\mathbf{x}_i$ in dataset, the CAE $f_{\text{CAE}}(\mathbf{x}_i, \mathbf{w})$ with model parameter $\mathbf{w}$ can be formulated as following:
\begin{equation}
\hat{\mathbf{x}_i} = f_{\text{CAE}}(\mathbf{x}_i, \mathbf{w}) = f_{\text{d}}( f_{\text{e}}(\mathbf{x}_i, \mathbf{w}_e), \mathbf{w}_d)
\end{equation}
where $\mathbf{w}_e$ and $\mathbf{w}_d$ is the parameters of the encoder and decoder, respectively. $\hat{\mathbf{x}_i}$ is the reconstructed data. To measure the reconstruction quality, the most commonly used loss function for anomaly detection is mean square error (MSE), i.e., $ L_{\text{MSE}} = \|\mathbf{x}_i - \hat{\mathbf{x}_i}\|_2^2$. Thus, the training objective of CAE can be constructed as:
\begin{equation}
\label{eq:spl_mad}
\underset{\mathbf{w}}{\text{min}} \sum_{i=1}^{n} L_{\text{MSE}}(f_{\text{CAE}}(\mathbf{x}_i, \mathbf{w}), \mathbf{x}_i).
\end{equation}

\textbf{Self-paced learning (SPL):} we introduce first the conventional SPL as preliminaries based on \cite{DBLP:conf/aaai/FanHLH17,DBLP:conf/nips/KumarPK10}.
Given a training dataset $\mathbf{D} = \{ (\mathbf{x}_i, y_i) \}_{i=1}^{n}$ with $n$ samples, where $\mathbf{x}_i \in R^d$ is the $i$-th sample and $y_i$ is the learning target (i.e., ground truth label). In our case, we use a fully convolutional autoencoder (CAE) to reconstruct the input image, that is, $y_i$ is equal to input $\mathbf{x}_i$. A learned model is denoted as $f(., \mathbf{w})$ where $\mathbf{w}$ is the model parameter. Let $L(f(\mathbf{x}_i, \mathbf{w}), y_i)$ denote the loss function that computes the cost between estimated data $f(\mathbf{x}_i, \mathbf{w})$ and target object $y_i$ of $i$-th sample. The sample weights is denoted as $\mathbf{v} = [v_1, ..., v_n]$. Then, the objective of SPL can be presented as a union of a weighted loss term on all samples and a general self-paced regularizer imposed on sample weights. The object of SPL can be expressed as the following minimization problem:
\begin{equation}
\label{eq:spl}
    \underset{\mathbf{w},\mathbf{v} \in [0,1]^n}{\text{min}} \: \mathbb{E}(\mathbf{w},\mathbf{v}; \lambda ) = \sum_{i=1}^{n} v_i L(f(\mathbf{x}_i, \mathbf{w}), y_i) + g(\lambda, v_i),
\end{equation}
where $g(\lambda, v_i)$ is the self-paced regularizer with a penalty parameter $\lambda$ that controls the learning space. Alternative search strategy (ASS) \cite{DBLP:conf/nips/KumarPK10} is generally used for solving Eq. \ref{eq:spl} that alternatively optimizes model weight $\mathbf{w}$ and sample weight $\mathbf{v}$ while keeping the other fixed. For example, given a fixed $\mathbf{v}$, the minimization over $\mathbf{w}$ is a weighted loss minimization problem which is independent of regularizer $g(\lambda, v_i)$. Through such a jointly learning process of $\mathbf{w}$ and $\mathbf{v}$ by ASS with gradually increasing the value of $\lambda$, more samples can be automatically included in the training process from easy to hard in a self-paced manner based on training losses.

\subsection{Unsupervised SPL-MAD}
\label{ssec:spl_mad}

\textbf{Reconstruction behaviour exploration:} In the context of anomaly detection, AE/CAE models trained on normal data are expected to produce higher reconstruction error for the abnormal data than normal data \cite{DBLP:conf/cvpr/0003CNRD16}. However, this assumption does not always hold as the reconstruction behavior of anomaly inputs is unclear when no anomalies exist in the training set. Zong \etal \cite{DBLP:conf/iclr/ZongSMCLCC18} observed that abnormal data obtains somehow lower reconstruction error than normal data. Hence, we will first analyze the reconstruction behavior in the MAD task by training a CAE on bona fides and limited morphed attacks. As discussed in Section \ref{sec:related_work},  the insufficient number of identities in the existing MAD datasets is one possible reason for the poor generalization of MAD performance. To address this issue, we utilize the in-the-wild CASIA-WebFace dataset \cite{casiawebface} as our assumed to be "mostly normal" samples, considering its diversity in capture environment, sensor, and identity. Specifically, the CASIA-WebFace dataset \cite{casiawebface} consists of 494,414 images across 10,575 identities collected from the web and is used for face verification and identification tasks. Then, we use an additional MAD dataset, namely SMDD \cite{damer2022privacy}, together with CASIA-WebFace for exploration of reconstruction behavior. We select the SMDD dataset due to its privacy-friendly property and diversity in identity. The detailed information of SMDD is presented later in Section \ref{ssec:datasets}. To investigate our conclusions hold on other datasets, we also did the same using the MorGAN-LMA (landmark-based morphs) dataset \cite{DBLP:conf/btas/DamerS0K18}.

Figure \ref{fig:loss_curves} (a) and (c) illustrate the average reconstruction errors on MorGAN-LMA and SMDD test set by a vanilla CAE model trained on the unlabeled combined dataset in each epoch. It can be clearly observed that the average reconstruction errors of attacks are lower than bona fides. Meanwhile, the error gap between bona fides and attacks consistently persists as the training continues. This finding is in stark contrast to the assumption \cite{DBLP:conf/cvpr/0003CNRD16} in general anomaly detection, indicating that morph attacks are easier to reconstruct. The possible reasons are: 1) the possible artefacts resulting from various morphing processes induce an ambiguity in some of the image details, which might make the image similar to a wider range of reconstructions, as they make it similar to faces of multiple identities, and thus result in a lower reconstruction error. 2) the blending artefacts existing in some morphed images might be easier to decode (less sharp information) from attack encoding. Overall, such feasible error gaps prove that the morphs can be detected in an unsupervised manner even if the model is trained on data, including polluted attack data.

\textbf{SPL-MAD:}
Despite the error gaps between bona fides and attacks observed in Figure \ref{fig:loss_curves}, the gaps become gradually smaller as the training continues. This is probably caused by the model learning anomalous patterns from the polluted attack samples as the training continues and thus leading to a degraded ability of the model to remove anomalies. To address this concern, we propose to incorporate the SPL paradigm into the training, aiming to continually remove the suspicious attacks in the training phase.
As we introduced before, the SPL paradigm consists of a problem-specific weighted term on all samples and an SPL regularizer on sample weights. Due to such ability of weight adjustment, SPL can enhance the robustness and generalizability of the model in polluted data. Therefore, we incorporate the SPL paradigm into our unsupervised MAD learning by assigning smaller weights to suspicious morphed attacks.
Our SPL-MAD solution is defined by:
\begin{equation}
\label{eq:our_spl_mad}
 \underset{\mathbf{w},\mathbf{v} \in [0,1]^n}{\text{min}} \sum_{i=1}^{n} v_i L_{\text{MSE}}(f_{\text{CAE}}(\mathbf{x}_i, \mathbf{w}), \mathbf{x}_i) + g(\lambda, v_i),
\end{equation}
where $f_{\text{CAE}}(\mathbf{x}_i, \mathbf{w})$ is the reconstructed image and $L_{\text{MSE}}(.)$ is the reconstruction loss (MSE in our case). The Eq. \ref{eq:our_spl_mad} is optimized by ASS. 
First, when sample weight $\mathbf{v} \in [0,1]^n$ in the regularizer is fixed, the minimization over $\mathbf{w}$ is a weighted loss minimization problem and the optimal model $\mathbf{w}^*$ is determined as:
\begin{equation}
\label{eq:w_op}
    \mathbf{w}^* = \underset{\mathbf{w}}{\text{argmin}} \sum_{i=1}^{n} v_i L_{\text{MSE}}(f_{\text{CAE}}(\mathbf{x}_i, \mathbf{w}), \mathbf{x}_i).
\end{equation}
The Eq. \ref{eq:w_op} is solved by gradient descent in the training phase. Alternatively, given model parameter $\mathbf{w}$, the optimal weight of the $i$-th sample $v_i^*$ is computed by:
\begin{equation}
\label{eq:v_op}
    v_i^* = \underset{v_i \in [0,1]}{\text{argmin}} \sum_{i=1}^{n} v_i L_{\text{MSE}}(f_{\text{CAE}}(\mathbf{x}_i, \mathbf{w}), \mathbf{x}_i) + g(\lambda, v_i).
\end{equation}
Based on the observation in Figure \ref{fig:loss_curves} that lower reconstruction losses are achieved by morphs than by bona fides at the beginning of training and a smaller difference between bona fides and attacks in the later epochs of training, we adapted general SPL rules to meet our needs: 1) $v_i^*$ is monotonically increasing w.r.t $\mathcal{L}_i$, the reconstruction loss of the $i$-th sample, which guides the model to select potential bona fide samples with larger losses in favor of suspicious morphs (morphs or images that have properties similar to a morphed image) with smaller losses. 2) $v_i^*$ is monotonically decreasing w.r.t $\lambda$, which means that a larger $\lambda$ has a higher tolerance to the losses (i.e., smaller sample weight) and can remove more suspicious samples. 3) $g(\lambda, v_i)$ is convex w.r.t $v_i \in [0,1]$ to ensure the soundness of SPL regularizer for optimization. Therefore, we selected a linear SPL regularizer proposed in \cite{DBLP:conf/mm/JiangMMH14} and modified its close-formed optimal solution formulation as:
\begin{equation}
\label{eq:optimal_v}
v_i^* = 
\begin{cases}
1 - \frac{\lambda}{\mathcal{L}_i}, & \mathcal{L}_i > \lambda  \\
0, &  \mathcal{L}_i \leq \lambda
\end{cases}
,
\end{equation}
Thus, Eq. \ref{eq:optimal_v} helps our unsupervised training process: the sample weight of data with a loss smaller than $\lambda$ (suspicious samples) is set to zero; the data with a large loss is assigned with a relatively large sample weight, which encourages CAE model to focus on the learning of potential of normal (assumed to be bona fide) samples. 

Furthermore, we determine a self-adaptive $\lambda$ by considering the reconstruction errors in each training step. We gradually increase the $\lambda$ to the maximum $\lambda_{max} = \mu(s) - \sigma(s)$ with the increasing training step $s$ by following the equation:
\begin{equation}
\label{eq:lambda}
    \lambda = \text{min}(\mu(s) - (m - r \cdot s) \cdot \sigma(s), \quad \lambda_{max})
\end{equation}
where $\mu(s)$ and $\sigma(s)$ denote the mean and standard deviation of all data in the current training step $s$. $m$ is the constant initial standard deviation range, and $r$ is the constant shrink rate. To be consistent with our modified SPL learning needs, $r$ is usually set to a small value, where only a few easy samples with very low reconstruction losses are removed at the early training stage. In our case, we assume that the majority of (or all) training samples are normal (assumed to be bona fide) samples. Therefore, the coefficient $(m - r \cdot s)$ of $\sigma(s)$ is limited to minimum 1 (i.e., $1 \cdot \sigma(s)$ in $\lambda_{max}$) to maintain the majority of the data. Our SPL-MAD paradigm is presented in Algorithm \ref{algo:spl_mad}.

\begin{algorithm}
    \SetKwInOut{KwIn}{Input}
    \SetKwInOut{KwOut}{Output}
    \KwIn{Input unlabeled data $\mathbf{D} = \{ \mathbf{x}_i \}_{i=1}^{n}$, a CAE model with parameters $\mathbf{w}$, training step size $s$}
    \KwOut{Updated model parameters $\mathbf{w}$}
    
    Initialize model parameter $\mathbf{w}$, sample weights $\mathbf{v}^*$ and $s=0$
    
    \Repeat{convergence}{
    Randomly sample a batch size of data from $\mathbf{D}$\;
    Calculate the reconstruction loss $L_{\text{MSE}}$ by forward propagation \;
    Compute $\lambda$ based on step-size $s$ by Eq. \ref{eq:lambda} \;
    $s = s + 1$ \;
    Updated sample weights $\mathbf{v}^*$ by Eq. \ref{eq:optimal_v} \;
    Update model parameters $\mathbf{w}^*$ by Eq. \ref{eq:w_op} \;
    }
    \KwRet{$\mathbf{w}^*$}
    \caption{The proposed SPL-MAD algorithm}
    \label{algo:spl_mad}
\end{algorithm}

\vspace{-5mm}
\section{Experimental setup} 
\subsection{Datasets}
\label{ssec:datasets}

To evaluate the generalizability of our MAD solutions on unknown attacks, we use eight publicly available MAD datasets: LMA-DRD (PS) \cite{DBLP:conf/isvc/DamerSFBKK21}, LMA-DRD (D) \cite{DBLP:conf/isvc/DamerSFBKK21}, MorGAN-LMA \cite{DBLP:conf/btas/DamerS0K18}, MorGAN-GAN \cite{DBLP:conf/btas/DamerS0K18}, FRLL-Morphs \cite{Sarkar2020}, FERET-Morphs \cite{Sarkar2020}, FRGC-Morphs \cite{Sarkar2020}, and one synthetic morphing attack detection development dataset (SMDD) \cite{damer2022privacy}. It should be noted that our model is trained on CASIA-WebFace dataset \cite{casiawebface} which is used for face recognition tasks, and does not contain any information on the images being manipulated or not (morphed, beautified, re-digitized, and so on.). Furthermore, a privacy-friendly SMDD training set is added to CASIA-WebFace to explore the effect of unlabeled attack contamination in the unsupervised training phase. 

\textbf{LMA-DRD (D) and LMA-DRD (PS):} The bona fide samples in LMA-DRD (D) and LMA-DRD (PS) are selected from VGGFace2 dataset \cite{DBLP:conf/fgr/CaoSXPZ18} and morphed attack samples are created by OpenCV morphing \cite{openCVmorph} and following parametrization in \cite{DBLP:conf/icb/RaghavendraRVB17} into two ways: "D" refers to digital and "PS" refers to re-digitized (print and scan). In our case, we only use the test set for unknown MAD, which consists of 123 bona fide and 88 morphs, for each D and PS.

\textbf{MorGAN-LMA and MorGAN-GAN:} The bona fide samples in MorGAN are selected from CelebA \cite{DBLP:conf/iccv/LiuLWT15} and morphs are created by either OpenCV landmark-based morphing \cite{openCVmorph} (denoted as MorGAN-LMA) or GAN-based solution presented in \cite{DBLP:conf/btas/DamerS0K18} (denoted as MorGAN-GAN). Each of the test sets contains 750 bona fides and 500 morphs. Note that the generated faces by the GAN-based solution in this dataset are of the relatively low resolution of $64 \times 64$ pixels.

 \textbf{FRLL-, FERET-, and FRGC-Morphs:} The morph samples in the FRLL-Morphs \cite{Sarkar2020} dataset is generated based on the publicly available Face Research London Lab dataset \cite{DeBruine_FRLL}. FRLL-Morphs contains five different morphing methods: OpenCV \cite{openCVmorph}, FaceMorpher \cite{Facemorpher}, StyleGAN2 \cite{DBLP:conf/nips/KarrasAHLLA20,DBLP:conf/iwbf/VenkateshZRRDB20}, and WebMorpher \cite{webmorph}, and AMSL \cite{amsl}. 
FRLL-Morphs is created for the evaluation of attack vulnerability and MAD performance (i.e., containing only a test set) and is thus suitable in our case for evaluating our solution. 
Each test set contains 204 bona fides and 1,222 attacks (per morphing type). 
The FERET-Morphs \cite{Sarkar2020} datasets contains bona fide images from FERET \cite{PHILLIPS1998295} and three types of morph attacks generated from OpenCV \cite{openCVmorph}, FaceMorpher \cite{Facemorpher}, and StyleGAN2 \cite{DBLP:conf/nips/KarrasAHLLA20,DBLP:conf/iwbf/VenkateshZRRDB20}. The test set contains 1,361 bona fides and between 523 and 525 attacks per morph type.
Similarly, the FRGC-Morphs \cite{Sarkar2020} datasets contains bona fide images from FRGC v2.0 \cite{DBLP:conf/cvpr/PhillipsFSBCHMMW05} and three types of morph attacks generated from OpenCV \cite{openCVmorph}, FaceMorpher \cite{Facemorpher}, and StyleGAN2 \cite{DBLP:conf/nips/KarrasAHLLA20,DBLP:conf/iwbf/VenkateshZRRDB20}. The test set contains 3,069 bona fides and between 961 and 963 attacks per morph type.

\textbf{SMDD:} The SMDD \cite{damer2022privacy} dataset is a synthetic-based MAD database that bona fide images are created by StyleGAN2-ADA \cite{DBLP:conf/nips/KarrasAHLLA20,DBLP:conf/iwbf/VenkateshZRRDB20} and morph attacks are generated based on such bona fide by using OpenCV morphing technique \cite{openCVmorph}. The training set and test set contain 25,000 bona fides and 15,000 morphs, each. Their extensive results showed that SMDD could be served as an effective training set for the MAD model, even when the MAD model encounters attacks created with unknown morphing techniques or data sources. Moreover, by considering the privacy-friendly characteristic and diversity of the SMDD dataset, we use the SMDD training set for further attack contamination exploration.

For the datasets that were also explored as the training data for supervised MAD methods in the experiments, only their respective train set was used for training, and only their test sets were used for evaluation. Both sets are identity-disjoint in all the datasets.

\subsection{Implementation details}
We use a CAE model consisting of seven convolutional blocks as our SPL-MAD backbone. We use a large-scale face dataset CASIA-WebFace \cite{casiawebface} for face verification and identification task to train the CAE model.
All the databases are cropped using landmark points obtained from Multi-task Cascaded Convolutional Networks (MTCNN) \cite{mtcnn}.
All crops are resized to $224 \times 224$ to match the network input size.
Overall, the input image size is $224 \times 224 \times 3$ for training. In the training phase, we use the stochastic gradient descent (SGD) optimizer with a momentum of 0.9 and a weight decay of 5e-4, and the initial learning rate is 1e-5. The degradation of the learning rate is controlled by an exponential learning scheduler with a gamma of 0.98. The batch size in our experiment is 64, and the number of training epochs is 25. The first five epoch is used for warm-up without SPL. The implementation is based on in PyTorch toolbox \cite{NEURIPS2019_9015}. The initial standard deviation range $m$ and the shrink rate $r$ are set to 4 and 5e-3, respectively. 

\subsection{Evaluation metrics}
We follow the standard definitions in ISO/IEC 30107-3 \cite{ISO301073} to evaluate MAD performance, that is, Attack Presentation Classification Error Rate (APCER) and Bona fide presentation Classification Error Rate (BPCER). APCER is the proportion of attack samples misclassified as bona fides, and BPCER is the proportion of bona fide samples misclassified as attacks. To report the overall MAD performance and for comparison with other well-established MAD solutions, we report the equal error rate (EER) value where APCER and BPCER are equal. Furthermore, we plot receiver operating characteristic (ROC) curves where the x-axis is APCER, and the y-axis is (1-BPCER) at different operation points to give a visual evaluation on a wider range.

\vspace{-3mm}
\section{Results}
In this section, we first analyze the experimental results of our developed SPL-MAD approach from two aspects: the contribution of the adapted SPL paradigm and the robustness of our approach on morphing attack contamination. Then, to put the performance of our unsupervised solution in the perspective of the supervised MAD performance, we compare our experimental results with three diverse supervised and well-performing MAD methods trained on five different sets of training data, totalling 15 supervised MAD solutions. 

\begin{table}[thbp!]
\begin{center}
\def\arraystretch{1.2}
\resizebox{0.49\textwidth}{!}{
\begin{tabular}{ll|cc|cc} 
\hline
\multicolumn{2}{c|}{\multirow{2}{*}{\diagbox[innerwidth=\textwidth*1/4]{Test data}{Train data}}} & \multicolumn{2}{c|}{CASIA-Web} & \multicolumn{2}{c}{CASIA-Web + SMDD} \\ \cline{3-6} 
\multicolumn{2}{c|}{} & \multicolumn{1}{c}{-} & SPL & \multicolumn{1}{c}{-  } & SPL \\ \hline 
\multicolumn{1}{l|}{\multirow{2}{*}{LMA-DRD}} & D & \multicolumn{1}{c}{31.81} & \textbf{18.18} & \multicolumn{1}{c}{22.73} & \textbf{20.45} \\ 
\multicolumn{1}{l|}{} & PS & \multicolumn{1}{c}{26.14} & \textbf{22.73} & \multicolumn{1}{c}{29.54} & 29.54 \\ \hline 
\multicolumn{1}{l|}{\multirow{2}{*}{MorGAN}} & LMA & \multicolumn{1}{c}{42.95} & \textbf{14.46} & \multicolumn{1}{c}{15.86} & \textbf{15.06} \\
\multicolumn{1}{l|}{} & GAN & \multicolumn{1}{c}{61.04} & \textbf{40.56} & \multicolumn{1}{c}{44.78} & \textbf{39.35} \\ \hline 
\multicolumn{1}{l|}{\multirow{5}{*}{FRLL-Morphs}} & OpenCV & \multicolumn{1}{c}{35.36} & \textbf{3.63} & \multicolumn{1}{c}{\textbf{4.71}} & 5.78 \\ 
\multicolumn{1}{l|}{} & FaceMorpher & \multicolumn{1}{c}{31.66} & \textbf{2.98} & \multicolumn{1}{c}{\textbf{2.87}} & 4.67 \\
\multicolumn{1}{l|}{} & StyleGAN2 & \multicolumn{1}{c}{34.69} & \textbf{15.14} & \multicolumn{1}{c}{\textbf{12.11}} & 12.92 \\ 
\multicolumn{1}{l|}{} & WebMorph & \multicolumn{1}{c}{37.51} & \textbf{12.29} & \multicolumn{1}{c}{\textbf{11.22}} & 15.72 \\
\multicolumn{1}{l|}{} & AMSL & \multicolumn{1}{c}{36.22} & \textbf{11.22} & \multicolumn{1}{c}{\textbf{8.87}} & 12.09 \\ \hline 
\multicolumn{1}{l|}{\multirow{3}{*}{FERET-Morphs}} & OpenCV & \multicolumn{1}{c}{45.24} & \textbf{32.13} & \multicolumn{1}{c}{36.14} & \textbf{30.21} \\ 
\multicolumn{1}{l|}{} & FaceMorpher & \multicolumn{1}{c}{39.81} & \textbf{27.69} & \multicolumn{1}{c}{36.14} & \textbf{25.76} \\
\multicolumn{1}{l|}{} & StyleGAN2 & \multicolumn{1}{c}{41.52} & \textbf{32.57} & \multicolumn{1}{c}{34.28} & \textbf{28.95} \\ \hline
\multicolumn{1}{l|}{\multirow{3}{*}{FRGC-Morphs}} & OpenCV & \multicolumn{1}{c}{48.23} & \textbf{36.11} & \multicolumn{1}{c}{21.62} & \textbf{19.54} \\ 
\multicolumn{1}{l|}{} & FaceMorpher & \multicolumn{1}{c}{47.24} & \textbf{23.99} & \multicolumn{1}{c}{19.67} & \textbf{18.42} \\
\multicolumn{1}{l|}{} & StyleGAN2 & \multicolumn{1}{c}{46.62} & \textbf{36.79} & \multicolumn{1}{c}{19.63} & \textbf{15.57} \\ \hline 
\multicolumn{2}{l|}{SMDD} & \multicolumn{1}{c}{29.59} & \textbf{7.60} & \multicolumn{1}{c}{\textbf{7.01}} & 9.19 \\ \hline \hline
\multicolumn{2}{l|}{Average performance} & \multicolumn{1}{c}{39.73} & \textbf{21.13} & \multicolumn{1}{c}{20.45} & \textbf{18.95} \\ \hline
\end{tabular}}
\caption{The MAD performance in terms of EER (\%) for the ablation study on SPL paradigm and on data contamination. The bold numbers indicate the lowest EER values in two training data protocols: face recognition dataset CASIA-WebFace \cite{casiawebface}, and CASIA-WebFace with a MAD dataset SMDD \cite{damer2022privacy}. '-' refers to the baseline model training without SPL paragiam, and 'SPL' refers to the training with SPL.}
\label{tab:ablation_results}
\end{center}
\vspace{-7mm}
\end{table}

\begin{figure*}[thbp!]
\begin{center}
\includegraphics[width=0.99\linewidth]{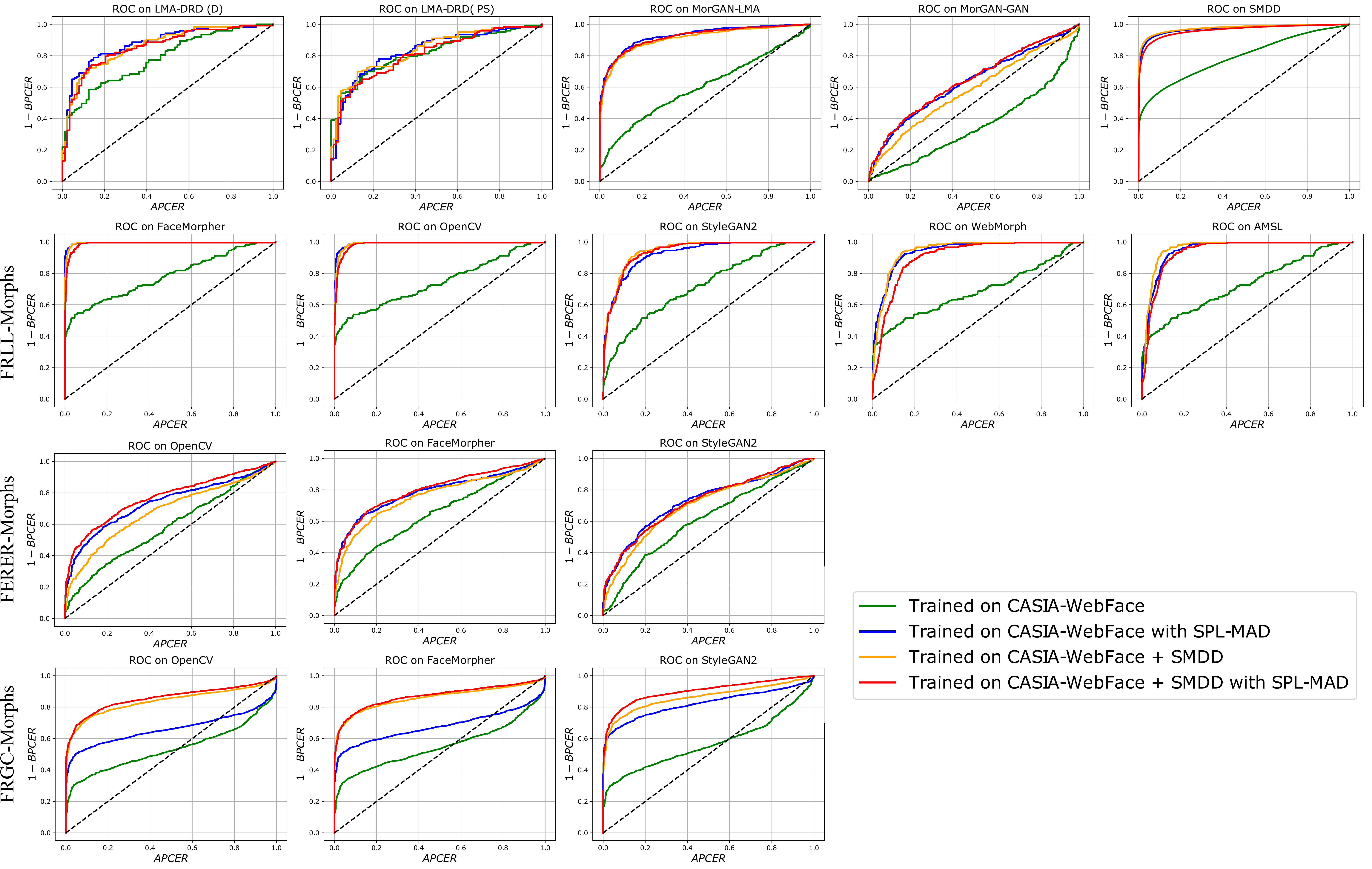}
\caption{ROC curves achieved on different test sets and four different training settings. In most cases, the curves indicate better MAD performance when including our modified SPL paradigm and when including data contamination by SMDD. }
\label{fig:roc}
\end{center}
\vspace{-7mm}
\end{figure*}

\subsection{Ablation study on SPL paradigm}
To illustrate the effectiveness of the SPL paradigm within our SPL-MAD, we trained an unsupervised CAE model without SPL (denoted as baseline) and with SPL (i.e., SPL-MAD), respectively. Table \ref{tab:ablation_results} shows the comparison results on various test sets. The lowest EER values on each training data protocol (i.e., CASIA-WebFace and CASIA-WebFace with SMDD \cite{damer2022privacy}) are shown in bold, respectively. From the results, we can observe that: 1) Including SPL in the training paradigm (i.e, SPL-MAD)  consistently outperformed the training without SPL when training on uncontaminated data. 2) Similarly, including SPL in the training paradigm also outperformed the model trained without SPL on most test sets when the training data was contaminated with the SMDD data. 3) Overall, a model trained with the SPL paradigm yields better average performance than without SPL. For example, EER is decreased from 39.73\% obtained by the baseline model to 21.13\% by SPL-MAD and decreased from 20.45\% to 18.95\% in two training data protocols, respectively. Such observations point out the significant contribution of the SPL paradigm to the improvement of the MAD performance. In addition to quantitative results, the larger reconstruction error gaps between bona fides and attacks in Figure \ref{fig:loss_curves} support our conclusion and our motivation behind using the SPL paradigm. As shown in Figure \ref{fig:loss_curves}, the SPL-MAD model achieved a larger gap between attack and bona fide than the baseline model on both MorGAN-LMA (error gap increases from 0.62 to 1.46) and SMDD dataset (error gap increases from 3.26 to 5.07). Furthermore, we plot the ROC curves for additional visual analysis. The ROC curves in Figure \ref{fig:roc} show that the blue and red curves of SPL-MAD model are placed above the green and yellow curves of the baseline models (no SPL) in most cases. Only for the FRLL-based attacks, and only in the case of contaminated training data, the benefit of the SPL is not as clear as both with SPL and without SPL training paradigms achieve very close performances. These graphical observations are consistent with the previous quantitative findings. In general, the SPL module plays an important role in our unsupervised MAD learning and enhances the overall improvement of MAD performance, as theorized earlier in Section \ref{sec:method}.

\subsection{Ablation study on data contamination}
To increase the diversity of bona fide identities and thus enhance the generalizability of MAD models, we leverage the publicly available face recognition datasets. However, most face recognition datasets were collected in the wild and might unintentionally include manipulated images. As a result, we provide two training data protocols by adding contaminated attacks to further demonstrate the robustness and effectiveness of our method trained purely on unlabeled data. One training set is a pure face recognition dataset CASIA-WebFace \cite{casiawebface}, and the other one is a combined training set consisting of the face recognition dataset and a face MAD training set (including both bona fides and attacks), that is CASIA-WebFace \cite{casiawebface} and SMDD \cite{damer2022privacy}, respectively. The ratio of bona fide to attack data in the combined dataset is around 35:1. We stress again that our unsupervised attack does not consider any labels during training, even for the contamination data. Table \ref{tab:ablation_results} shows the comparison results in terms of the EER value obtained by the model trained on both training data protocols. It can be observed that baseline and SPL-MAD model trained with contaminated data gains 19.33\% and 2.187\% overall performance improvement over models trained on uncontaminated CASIA-WebFace. The reason behind such improvement might be that more data is included in the training phase, and the majority of training data are still bona fides. This observation suggests that our unsupervised MAD method sees a steady rise in the performance irrespective of the attack contamination. 
Moreover, the ROC curves in Figure \ref{fig:roc} confirms our quantitative findings and indicate that our unsupervised model proves to be effective even under data contamination scenario in most cases.

\begin{table*}[thbp!]
\begin{center}
\def\arraystretch{1.2}
\resizebox{0.99\textwidth}{!}{
\begin{tabular}{ll|ccccc|ccccc|ccccc|c}
\hline
\multicolumn{2}{c|}{\multirow{3}{*}{\diagbox[innerwidth=\textwidth*1/4]{Test data}{Train data}}} & \multicolumn{15}{c|}{Supervised} & \multicolumn{1}{c}{Unsupervised} \\ \cline{3-18} 
\multicolumn{2}{c|}{} & \multicolumn{5}{c|}{MixFaceNet} & \multicolumn{5}{c|}{PW-MAD} & \multicolumn{5}{c|}{Inception} & \multicolumn{1}{c}{SPL-MAD} \\
\multicolumn{2}{c|}{} & \multicolumn{1}{c}{D} & \multicolumn{1}{c}{PS} & \multicolumn{1}{c}{-LMA} & \multicolumn{1}{c}{-GAN} & \multicolumn{1}{c|}{SMDD} & \multicolumn{1}{c}{D} & \multicolumn{1}{c}{PS} & \multicolumn{1}{c}{-LMA} & \multicolumn{1}{c}{-GAN} & \multicolumn{1}{c|}{SMDD} & \multicolumn{1}{c}{D} & \multicolumn{1}{c}{PS} & \multicolumn{1}{c}{-LMA} & \multicolumn{1}{c}{-GAN} & \multicolumn{1}{c|}{SMDD} & \multicolumn{1}{c}{(Our)} \\ \hline \hline
\multicolumn{1}{l|}{\multirow{2}{*}{LMAD-DRD}} & D & 15.68$^*$ & 18.03 & 17.06 & 25.01 & 19.42 & {20.8$^*$ } & 25.1 & 22.34 & 40.21 & 17.06 & {7.64$^*$ } & 17.06 & 15.68 & 50.77 & 15.11 & 20.45 \\
\multicolumn{1}{l|}{} & PS & 21.77 & {18.44$^*$ } & 27.05 & 27.05 & 23.72 & 26.48 & {23.72$^*$ } & 29.41 & 44.11 & 20.39 & 11.37 & {12.75$^*$ } & 22.34 & 38.42 & 19.01 & 29.54 \\ \hline 
\multicolumn{1}{l|}{\multirow{2}{*}{MorGAN}} & LMA & 39.42 & 22.89 & {10.61$^*$ } & 46.42 & 30.12 & 34.2 & 34.14 & {9.71$^*$ } & 34.37 & 27.31 & 38.55 & 31.73 & {8.43$^*$ } & 40.16 & 28.51 & 15.06 \\ 
\multicolumn{1}{l|}{} & GAN & 53.01 & 50.44 & 42.57 & {24.9$^*$ } & 42.64 & 52.04 & 46.59 & 42.8 & {8.84$^*$ } & 43.78 & 50.84 & 38.79 & 27.41 & {0.4$^*$ } & 44.34 & 39.35 \\ \hline
\multicolumn{1}{l|}{\multirow{5}{*}{FRLL-Morphs}} & OpenCV & 8.82 & 13.22 & 8.91 & 17.66 & 4.39 & 17.33 & 15.69 & 13.96 & 45.59 & 2.42 & 13.72 & 10.76 & 6.86 & 55.89 & 5.38 & 5.78 \\ 
\multicolumn{1}{l|}{} & FaceMorpher & 7.80 & 10.97 & 7.34 & 15.65 & 3.87 & 13.88 & 15.14 & 10.92 & 44.57 & 2.20 & 16.62 & 15.81 & 6.32 & 66.14 & 3.17 & 4.67 \\ 
\multicolumn{1}{l|}{} & StyleGAN2 & 20.07 & 15.29 & 13.41 & 23.51 & 8.89 & 29.97 & 27.64 & 18.11 & 48.53 & 16.64 & 37.24 & 19.58 & 20.56 & 55.03 & 11.37 & 12.92 \\ 
\multicolumn{1}{l|}{} & WebMorph & 25.97 & 29.04 & 20.61 & 30.39 & 12.35 & 33.78 & 28.51 & 35.75 & 52.43 & 16.65 & 57.38 & 58.32 & 30.88 & 77.42 & 9.86 & 15.72 \\ 
\multicolumn{1}{l|}{} & AMSL & 24.53 & 27.59 & 19.24 & 30.03 & 15.18 & 36.25 & 32.95 & 34.38 & 48.52 & 15.18 & 49.02 & 61.44 & 9.80 & 86.49 & 10.79 & 12.09 \\\hline
\multicolumn{1}{l|}{\multirow{3}{*}{FERET-Morphs}} & OpenCV & 28.12 & 32.19 & 31.57 & 33.86 & 31.74 & 37.27 & 45.29 & 34.27 & 43.11 & 39.93 & 6.39 & 7.23 & 42.12 & 13.62 & 59.32 & 30.21 \\ 
\multicolumn{1}{l|}{} & FaceMorpher & 22.57 & 29.48 & 27.9 & 31.81 & 23.69 & 35.16 & 44.3 & 28.24 & 40.4 & 29.41 & 5.17 & 6.91 & 36.53 & 18.36 & 46.94 & 25.76 \\ 
\multicolumn{1}{l|}{} & StyleGAN2 & 29.57 & 29.02 & 35.46 & 39.41 & 39.85 & 44.25 & 45.3 & 29.7 & 42.47 & 47.2 & 9.03 & 7.12 & 35.29 & 15.09 & 60.05 & 28.95 \\ \hline
\multicolumn{1}{l|}{\multirow{3}{*}{FRGC-Morphs}} & OpenCV & 23.81 & 25.04 & 31.62 & 21.11 & 20.67 & 57.06 & 48.6 & 29.74 & 53.55 & 26.45 & 34.32 & 13.65 & 36.17 & 59.66 & 19.63 & 19.54 \\ 
\multicolumn{1}{l|}{} & FaceMorpher & 22.83 & 23.54 & 29.38 & 19.98 & 18.10 & 56 & 50.7 & 30.49 & 51.61 & 23.4 & 34.96 & 19.71 & 35.1 & 56.91 & 16.06 & 18.42 \\ 
\multicolumn{1}{l|}{} & StyleGAN2 & 32.71 & 28.68 & 21.7 & 21.95 & 11.62 & 37.38 & 38.42 & 16.43 & 26.62 & 14.32 & 41.14 & 25.85 & 36.19 & 47.03 & 15.26 & 15.57 \\\hline
\multicolumn{2}{l|}{SMDD} & 10.34 & 9.26 & 5.11 & 11.69 & {2.51$^*$ } & 15.7 & 13.45 & 7.82 & 36.25 & {0.79$^*$ } & 6.42 & 21.88 & 12.49 & 38.38 & {0.42$^*$ } & 9.19 \\ \hline \hline
\multicolumn{2}{l|}{Average performance} & 24.76 & 24.31 & 22.60 & 26.37 & 20.42 & 35.12 & 34.12 & 25.62 & 43.49 & 22.82 & 27.48 & 23.72 & 24.92 & 49.17 & 24.32 & \textbf{18.95} \\ \hline
\end{tabular}}
\caption{The MAD performance in terms of EER (\%) in comparison with results of three supervise-learning-based and well-performing MAD solutions: MixFaceNet \cite{DBLP:conf/icb/BoutrosDFKK21}, PW-MAD \cite{DBLP:conf/isvc/DamerSFBKK21}, Inception \cite{DBLP:conf/cvip/RamachandraVRB18} trained on five datasets. The results under intra-dataset evaluation are marked with $^*$, and the best MAD performance among 16 MAD models is denoted in bold. Note that for a fair comparison, the intra-dataset results are neglected when calculating the average performance, and our unsupervised SPL-MAD results are reported under the data contamination scenario.}
\label{tab:comparison_results}
\end{center}
\vspace{-5mm}
\end{table*}

\begin{figure}[thbp!]
\begin{center}
\includegraphics[width=1.0\linewidth]{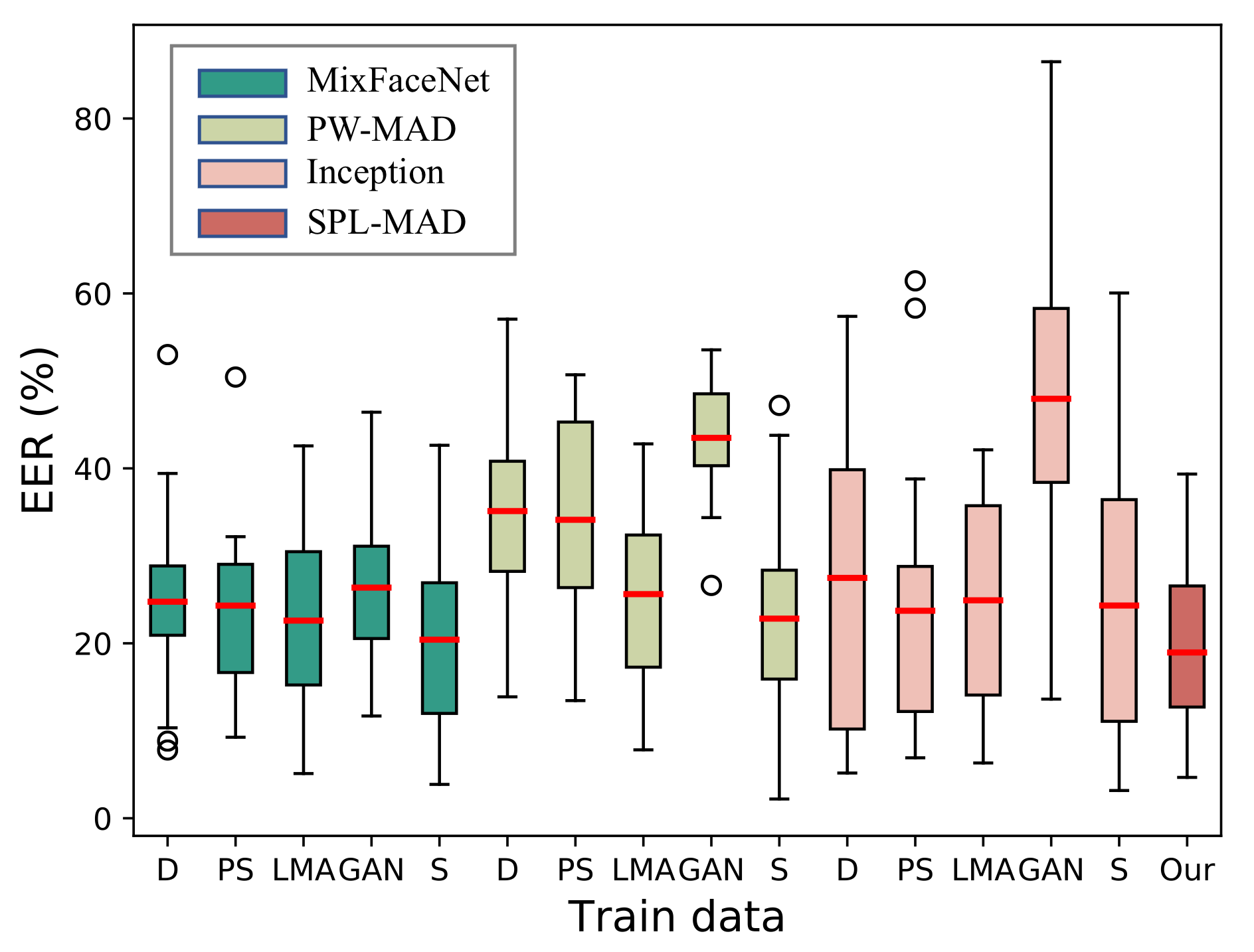}
\caption{The box and whisker plot of the performance variation per each model. Each box represents a model trained on different datasets. The x-axis is the trained dataset (D, PS, LMA, GAN, and S represent LMA-DRD (D), LMA-DRD (PS), MorGAN-LMA, MorGAN-GAN, SMDD, and an "our" unsupervised SPL-MAD trained on a combined set of CASIA-WebFace and SMDD). Four colors indicate the used MAD models, i.e., MixFaceNet, PW-MAD, Inception, and our proposed SPL-MAD. The y-axis is the EER (\%) value, and the red line within the box refers to the mean value of each model on all test sets (as in table \ref{tab:comparison_results}), the box range represents the standard deviation, and the "o" signs are extreme outliers. Note the low average EER value of our SPL-MAD, along with its low deviation and lack of extreme outlier performances.}
\vspace{-9mm}
\label{fig:performance_deviation}
\end{center}
\end{figure}

\vspace{-3mm}
\subsection{Comparison to supervised MAD}
Furthermore, to put our unsupervised SPL-MAD in the perspective of the supervised MAD performances, Table \ref{tab:comparison_results} presents the performance comparison of our method (i.e., SPL-MAD trained on contaminated data) to other three supervised single image MAD solutions \cite{damer2022privacy,DBLP:conf/isvc/DamerSFBKK21, DBLP:conf/cvip/RamachandraVRB18}. The results under the intra-dataset scenario are marked with $^*$ (which is expected to perform unfairly good), and the best average performance (lowest EER) among all models is in bold. It should be noted that we only consider the results in the cross-dataset setting when calculating the average performance for a fair comparison. From the results, we can conclude that our unsupervised SPL-MAD achieved comparable results with supervised methods in most test set cases. When considering the average performance on all datasets, our method obtains the lowest EER value (18.95\%), where the second-lowest EER is 20.42\%. The improvement in MAD performance can be attributed to the large-scale training data enabled by the unsupervised nature of the approach and the adaption of the SPL paradigm.
In Figure \ref{fig:performance_deviation}, we plot the mean and standard deviation of the EER values achieved by our proposed SPL-MAD, and the reported 15 supervised MADs on all the testing datasets (excluding intra-dataset tests). The plot stresses that the average MAD performance of the proposed approach is comparable and better than the supervised methods. However, more importantly, as expected from an unsupervised approach, it performs more consistently (low deviation with no extreme outliers) than supervised methods that commonly fail when facing unknown morphing attacks.

\vspace{-3mm}
\section{Conclusion}
In this work, we proposed a novel completely unsupervised MAD solution via self-paced anomaly detection (SPL-MAD), which benefits from the unsupervised nature of autoencoders training and the property of self-paced learning that automatically evaluates the training data without any prior knowledge. 
First, to address the lack of diversity and quantity of MAD datasets, we leverage the existing large-scale face recognition datasets to train our unsupervised model. In addition, we build a solid theoretical foundation for designing our unsupervised solution by empirically analyzing the reconstruction behavior on MAD data. 
This also led us to propose integrating our adaptive self-learning paradigm to improve the separability of reconstruction errors between bona fide and attack samples in a fully unsupervised manner.
The experimental results demonstrated on a diverse set of MAD data indicated the higher generalizability of our unsupervised SPL-MAD solution in comparison to a wide range of supervised MAD solutions in dealing with morphing attacks created from unknown morphing techniques or data sources.
\vspace{-5mm}
\paragraph{Acknowledgment:}
This research work has been funded by the German Federal Ministry of Education and Research and the Hessen State Ministry for Higher Education, Research and the Arts within their joint support of the National Research Center for Applied Cybersecurity ATHENE.

{\small
\bibliographystyle{ieee}
\bibliography{egbib}

\begin{thebibliography}{10}\itemsep=-1pt

\bibitem{DBLP:conf/icml/BengioLCW09}
Y.~Bengio, J.~Louradour, R.~Collobert, and J.~Weston.
\newblock Curriculum learning.
\newblock In {\em Proceedings of the 26th Annual International Conference on
  Machine Learning, {ICML} 2009, Montreal, Quebec, Canada, June 14-18, 2009},
  pages 41--48, 2009.

\bibitem{DBLP:conf/icb/BoutrosDFKK21}
F.~Boutros, N.~Damer, M.~Fang, F.~Kirchbuchner, and A.~Kuijper.
\newblock Mixfacenets: Extremely efficient face recognition networks.
\newblock In {\em International {IEEE} Joint Conference on Biometrics, {IJCB}
  2021, Shenzhen, China, August 4-7, 2021}, pages 1--8. {IEEE}, 2021.

\bibitem{boutros2021elasticface}
F.~Boutros, N.~Damer, F.~Kirchbuchner, and A.~Kuijper.
\newblock Elasticface: Elastic margin loss for deep face recognition.
\newblock In {\em {IEEE} Conference on Computer Vision and Pattern Recognition
  Workshops, {CVPR} Workshops 2022}, page~1, 2022.

\bibitem{DBLP:conf/fgr/CaoSXPZ18}
Q.~Cao, L.~Shen, W.~Xie, O.~M. Parkhi, and A.~Zisserman.
\newblock Vggface2: {A} dataset for recognising faces across pose and age.
\newblock In {\em 13th {IEEE} International Conference on Automatic Face {\&}
  Gesture Recognition, {FG} 2018, Xi'an, China, May 15-19, 2018}, pages 67--74.
  {IEEE} Computer Society, 2018.

\bibitem{DBLP:conf/dagm/DamerBWBTBK18}
N.~Damer, V.~Boller, Y.~Wainakh, F.~Boutros, P.~Terh{\"{o}}rst, A.~Braun, and
  A.~Kuijper.
\newblock Detecting face morphing attacks by analyzing the directed distances
  of facial landmarks shifts.
\newblock In {\em Pattern Recognition - 40th German Conference, {GCPR} 2018,
  Stuttgart, Germany, October 9-12, 2018, Proceedings}, volume 11269 of {\em
  Lecture Notes in Computer Science}, pages 518--534. Springer, 2018.

\bibitem{DBLP:conf/btas/DamerBSKK19}
N.~Damer, F.~Boutros, A.~M. Saladie, F.~Kirchbuchner, and A.~Kuijper.
\newblock Realistic dreams: Cascaded enhancement of gan-generated images with
  an example in face morphing attacks.
\newblock In {\em {BTAS}}, pages 1--10. {IEEE}, 2019.

\bibitem{DBLP:conf/bmvc/DamerD16}
N.~Damer and K.~Dimitrov.
\newblock Practical view on face presentation attack detection.
\newblock In {\em {BMVC}}. {BMVA} Press, 2016.

\bibitem{DBLP:conf/btas/DamerGZKK19}
N.~Damer, J.~H. Grebe, S.~Zienert, F.~Kirchbuchner, and A.~Kuijper.
\newblock On the generalization of detecting face morphing attacks as
  anomalies: Novelty vs. outlier detection.
\newblock In {\em {BTAS}}, pages 1--5. {IEEE}, 2019.

\bibitem{damer2022privacy}
N.~Damer, C.~A.~F. L{\'o}pez, M.~Fang, N.~Spiller, M.~V. Pham, and F.~Boutros.
\newblock Privacy-friendly synthetic data for the development of face morphing
  attack detectors.
\newblock In {\em {IEEE} Conference on Computer Vision and Pattern Recognition
  Workshops, {CVPR} Workshops 2022}, page~1, 2022.

\bibitem{ReGenM}
N.~Damer, K.~B. Raja, M.~S{\"{u}}{\ss}milch, S.~Venkatesh, F.~Boutros, M.~Fang,
  F.~Kirchbuchner, R.~Ramachandra, and A.~Kuijper.
\newblock Regenmorph: Visibly realistic {GAN} generated face morphing attacks
  by attack re-generation.
\newblock In {\em Advances in Visual Computing - 16th International Symposium,
  {ISVC} 2021, Virtual Event, October 4-6, 2021, Proceedings, Part {I}}, volume
  13017 of {\em Lecture Notes in Computer Science}, pages 251--264. Springer,
  2021.

\bibitem{DBLP:conf/btas/DamerS0K18}
N.~Damer, A.~M. Saladie, A.~Braun, and A.~Kuijper.
\newblock {MorGAN}: Recognition vulnerability and attack detectability of face
  morphing attacks created by generative adversarial network.
\newblock In {\em {BTAS}}, pages 1--10. {IEEE}, 2018.

\bibitem{DBLP:conf/icb/DamerSZWTKK19}
N.~Damer, A.~M. Saladie, S.~Zienert, Y.~Wainakh, P.~Terh{\"{o}}rst,
  F.~Kirchbuchner, and A.~Kuijper.
\newblock To detect or not to detect: The right faces to morph.
\newblock In {\em {ICB}}, pages 1--8. {IEEE}, 2019.

\bibitem{DBLP:conf/isvc/DamerSFBKK21}
N.~Damer, N.~Spiller, M.~Fang, F.~Boutros, F.~Kirchbuchner, and A.~Kuijper.
\newblock {PW-MAD:} pixel-wise supervision for generalized face morphing attack
  detection.
\newblock In {\em Advances in Visual Computing - 16th International Symposium,
  {ISVC} 2021, Virtual Event, October 4-6, 2021, Proceedings, Part {I}}, volume
  13017 of {\em Lecture Notes in Computer Science}, pages 291--304. Springer,
  2021.

\bibitem{DBLP:conf/btas/DamerWBBT0K18}
N.~Damer, Y.~Wainakh, V.~Boller, S.~von~den Berken, P.~Terh{\"{o}}rst,
  A.~Braun, and A.~Kuijper.
\newblock Crazyfaces: Unassisted circumvention of watchlist face
  identification.
\newblock In {\em {BTAS}}, pages 1--9. {IEEE}, 2018.

\bibitem{DBLP:conf/fusion/DamerZWSKK19}
N.~Damer, S.~Zienert, Y.~Wainakh, A.~M. Saladie, F.~Kirchbuchner, and
  A.~Kuijper.
\newblock A multi-detector solution towards an accurate and generalized
  detection of face morphing attacks.
\newblock In {\em 22th International Conference on Information Fusion, {FUSION}
  2019, Ottawa, ON, Canada, July 2-5, 2019}, pages 1--8. {IEEE}, 2019.

\bibitem{DBLP:conf/iciap/DebiasiDSRSBKU19}
L.~Debiasi, N.~Damer, A.~M. Saladie, C.~Rathgeb, U.~Scherhag, C.~Busch,
  F.~Kirchbuchner, and A.~Uhl.
\newblock On the detection of gan-based face morphs using established morph
  detectors.
\newblock In {\em Image Analysis and Processing - {ICIAP} 2019 - 20th
  International Conference, Trento, Italy, September 9-13, 2019, Proceedings,
  Part {II}}, volume 11752 of {\em Lecture Notes in Computer Science}, pages
  345--356. Springer, 2019.

\bibitem{webmorph}
L.~DeBruine.
\newblock debruine/webmorph: Beta release 2.
\newblock Jan. 2018.

\bibitem{DeBruine_FRLL}
L.~DeBruine and B.~Jones.
\newblock Face research lab london set, May 2017.

\bibitem{DBLP:conf/cvpr/DengGXZ19}
J.~Deng, J.~Guo, N.~Xue, and S.~Zafeiriou.
\newblock Arcface: Additive angular margin loss for deep face recognition.
\newblock In {\em {CVPR}}, pages 4690--4699. Computer Vision Foundation /
  {IEEE}, 2019.

\bibitem{DBLP:conf/aaai/FanHLH17}
Y.~Fan, R.~He, J.~Liang, and B.~Hu.
\newblock Self-paced learning: An implicit regularization perspective.
\newblock In {\em Proceedings of the Thirty-First {AAAI} Conference on
  Artificial Intelligence, February 4-9, 2017, San Francisco, California,
  {USA}}, pages 1877--1883, 2017.

\bibitem{DBLP:journals/pr/FangDKK22}
M.~Fang, N.~Damer, F.~Kirchbuchner, and A.~Kuijper.
\newblock Real masks and spoof faces: On the masked face presentation attack
  detection.
\newblock {\em Pattern Recognit.}, 123:108398, 2022.

\bibitem{DBLP:books/sp/16/FerraraFM16}
M.~Ferrara, A.~Franco, and D.~Maltoni.
\newblock On the effects of image alterations on face recognition accuracy.
\newblock In T.~Bourlai, editor, {\em Face Recognition Across the Imaging
  Spectrum}, pages 195--222. Springer, 2016.

\bibitem{MADVgg}
M.~Ferrara, A.~Franco, and D.~Maltoni.
\newblock Face morphing detection in the presence of printing/scanning and
  heterogeneous image sources.
\newblock {\em IET Biometrics}, 10(3):290--303, 2021.

\bibitem{DBLP:conf/cvpr/0003CNRD16}
M.~Hasan, J.~Choi, J.~Neumann, A.~K. Roy{-}Chowdhury, and L.~S. Davis.
\newblock Learning temporal regularity in video sequences.
\newblock In {\em 2016 {IEEE} Conference on Computer Vision and Pattern
  Recognition, {CVPR} 2016, Las Vegas, NV, USA, June 27-30, 2016}, pages
  733--742, 2016.

\bibitem{Ibsen2021}
M.~Ibsen, L.~J. Gonzalez-Soler, C.~Rathgeb, P.~Drozdowski, M.~Gomez-Barrero,
  and C.~Busch.
\newblock Differential anomaly detection for facial images.
\newblock In {\em 2021 IEEE International Workshop on Information Forensics and
  Security (WIFS)}, pages 1--6, 2021.

\bibitem{ICAO}
{International Civil Aviation Organization, ICAO}.
\newblock {Machine readable passports – part 9 – deployment of biometric
  identification and electronic storage of data in eMRTDs}.
\newblock {\em {Civil Aviation Organization (ICAO)}}, 2015.

\bibitem{ISO301073}
{International Organization for Standardization}.
\newblock {ISO/IEC DIS 30107-3:2016: Information Technology – Biometric
  presentation attack detection – P. 3: Testing and reporting}, 2017.

\bibitem{DBLP:conf/mm/JiangMMH14}
L.~Jiang, D.~Meng, T.~Mitamura, and A.~G. Hauptmann.
\newblock Easy samples first: Self-paced reranking for zero-example multimedia
  search.
\newblock In {\em Proceedings of the {ACM} International Conference on
  Multimedia, {MM} '14, Orlando, FL, USA, November 03 - 07, 2014}, pages
  547--556. {ACM}, 2014.

\bibitem{DBLP:conf/nips/KarrasAHLLA20}
T.~Karras, M.~Aittala, J.~Hellsten, S.~Laine, J.~Lehtinen, and T.~Aila.
\newblock Training generative adversarial networks with limited data.
\newblock In {\em Advances in Neural Information Processing Systems 33: Annual
  Conference on Neural Information Processing Systems 2020, NeurIPS 2020,
  December 6-12, 2020, virtual}, 2020.

\bibitem{DBLP:conf/nips/KumarPK10}
M.~P. Kumar, B.~Packer, and D.~Koller.
\newblock Self-paced learning for latent variable models.
\newblock In {\em Advances in Neural Information Processing Systems 23: 24th
  Annual Conference on Neural Information Processing Systems 2010. Proceedings
  of a meeting held 6-9 December 2010, Vancouver, British Columbia, Canada},
  pages 1189--1197, 2010.

\bibitem{DBLP:conf/iccv/LiuLWT15}
Z.~Liu, P.~Luo, X.~Wang, and X.~Tang.
\newblock Deep learning face attributes in the wild.
\newblock In {\em 2015 {IEEE} International Conference on Computer Vision,
  {ICCV} 2015, Santiago, Chile, December 7-13, 2015}, pages 3730--3738. {IEEE}
  Computer Society, 2015.

\bibitem{DBLP:conf/visapp/MakrushinND17}
A.~Makrushin, T.~Neubert, and J.~Dittmann.
\newblock Automatic generation and detection of visually faultless facial
  morphs.
\newblock In {\em {VISIGRAPP} {(6:} {VISAPP)}}, pages 39--50. SciTePress, 2017.

\bibitem{openCVmorph}
S.~Mallick.
\newblock Face morph using opencv — c++ / python.
\newblock {\em LearnOpenCV}, 1(1), 2016.

\bibitem{DBLP:journals/cviu/MassoliCAF21}
F.~V. Massoli, F.~Carrara, G.~Amato, and F.~Falchi.
\newblock Detection of face recognition adversarial attacks.
\newblock {\em Comput. Vis. Image Underst.}, 202:103103, 2021.

\bibitem{amsl}
T.~Neubert, A.~Makrushin, M.~Hildebrandt, C.~Kraetzer, and J.~Dittmann.
\newblock Extended stirtrace benchmarking of biometric and forensic qualities
  of morphed face images.
\newblock {\em IET Biometrics}, 7(4):325--332, 2018.

\bibitem{NEURIPS2019_9015}
A.~Paszke, S.~Gross, F.~Massa, A.~Lerer, J.~Bradbury, G.~Chanan, T.~Killeen,
  Z.~Lin, N.~Gimelshein, L.~Antiga, A.~Desmaison, A.~Kopf, E.~Yang, Z.~DeVito,
  M.~Raison, A.~Tejani, S.~Chilamkurthy, B.~Steiner, L.~Fang, J.~Bai, and
  S.~Chintala.
\newblock Pytorch: An imperative style, high-performance deep learning library.
\newblock In H.~Wallach, H.~Larochelle, A.~Beygelzimer, F.~d\textquotesingle
  Alch\'{e}-Buc, E.~Fox, and R.~Garnett, editors, {\em Advances in Neural
  Information Processing Systems 32}, pages 8024--8035. Curran Associates,
  Inc., 2019.

\bibitem{PHILLIPS1998295}
P.~Phillips, H.~Wechsler, J.~Huang, and P.~J. Rauss.
\newblock The feret database and evaluation procedure for face-recognition
  algorithms.
\newblock In {\em Image and Vision Computing}, volume~16, pages 295--306, 1998.

\bibitem{DBLP:conf/cvpr/PhillipsFSBCHMMW05}
P.~J. Phillips, P.~J. Flynn, W.~T. Scruggs, K.~W. Bowyer, J.~Chang, K.~Hoffman,
  J.~Marques, J.~Min, and W.~J. Worek.
\newblock Overview of the face recognition grand challenge.
\newblock In {\em 2005 {IEEE} Computer Society Conference on Computer Vision
  and Pattern Recognition {(CVPR} 2005), 20-26 June 2005, San Diego, CA,
  {USA}}, pages 947--954, 2005.

\bibitem{Facemorpher}
A.~Quek.
\newblock Facemorpher.
\newblock 2019.

\bibitem{DBLP:conf/btas/RaghavendraRB16}
R.~Raghavendra, K.~B. Raja, and C.~Busch.
\newblock Detecting morphed face images.
\newblock In {\em {BTAS}}, pages 1--7. {IEEE}, 2016.

\bibitem{DBLP:conf/icb/RaghavendraRVB17}
R.~Raghavendra, K.~B. Raja, S.~Venkatesh, and C.~Busch.
\newblock Face morphing versus face averaging: Vulnerability and detection.
\newblock In {\em {IJCB}}, pages 555--563. {IEEE}, 2017.

\bibitem{DBLP:conf/cvpr/RaghavendraRVB17a}
R.~Raghavendra, K.~B. Raja, S.~Venkatesh, and C.~Busch.
\newblock Transferable deep-cnn features for detecting digital and
  print-scanned morphed face images.
\newblock In {\em 2017 {IEEE} Conference on Computer Vision and Pattern
  Recognition Workshops, {CVPR} Workshops 2017, Honolulu, HI, USA, July 21-26,
  2017}, pages 1822--1830. {IEEE} Computer Society, 2017.

\bibitem{DBLP:conf/cvip/RamachandraVRB18}
R.~Ramachandra, S.~Venkatesh, K.~B. Raja, and C.~Busch.
\newblock Detecting face morphing attacks with collaborative representation of
  steerable features.
\newblock In {\em {CVIP} {(1)}}, volume 1022 of {\em AISC}, pages 255--265.
  Springer, 2018.

\bibitem{Sarkar2020}
E.~Sarkar, P.~Korshunov, L.~Colbois, and S.~Marcel.
\newblock Vulnerability analysis of face morphing attacks from landmarks and
  generative adversarial networks.
\newblock {\em arXiv preprint}, Oct. 2020.

\bibitem{DBLP:conf/iwbf/ScherhagRRGRB17}
U.~Scherhag, R.~Raghavendra, K.~B. Raja, M.~Gomez{-}Barrero, C.~Rathgeb, and
  C.~Busch.
\newblock On the vulnerability of face recognition systems towards morphed face
  attacks.
\newblock In {\em {IWBF}}, pages 1--6. {IEEE}, 2017.

\bibitem{DBLP:conf/iwbf/ScherhagRB18}
U.~Scherhag, C.~Rathgeb, and C.~Busch.
\newblock Performance variation of morphed face image detection algorithms
  across different datasets.
\newblock In {\em {IWBF}}, pages 1--6. {IEEE}, 2018.

\bibitem{soleymani2021mutual}
S.~Soleymani, A.~Dabouei, F.~Taherkhani, J.~Dawson, and N.~M. Nasrabadi.
\newblock Mutual information maximization on disentangled representations for
  differential morph detection.
\newblock In {\em Proceedings of the IEEE/CVF Winter Conference on Applications
  of Computer Vision}, pages 1731--1741, 2021.

\bibitem{DBLP:conf/wacv/SoleymaniDTDN21}
S.~Soleymani, A.~Dabouei, F.~Taherkhani, J.~M. Dawson, and N.~M. Nasrabadi.
\newblock Mutual information maximization on disentangled representations for
  differential morph detection.
\newblock In {\em {IEEE} Winter Conference on Applications of Computer Vision,
  {WACV} 2021, Waikoloa, HI, USA, January 3-8, 2021}, pages 1730--1740. {IEEE},
  2021.

\bibitem{DBLP:conf/iwbf/VenkateshZRRDB20}
S.~Venkatesh, H.~Zhang, R.~Ramachandra, K.~B. Raja, N.~Damer, and C.~Busch.
\newblock Can {GAN} generated morphs threaten face recognition systems equally
  as landmark based morphs? - vulnerability and detection.
\newblock In {\em 8th International Workshop on Biometrics and Forensics,
  {IWBF} 2020, Porto, Portugal, April 29-30, 2020}, pages 1--6. {IEEE}, 2020.

\bibitem{gdpr}
P.~Voigt and A.~v.~d. Bussche.
\newblock {\em The EU General Data Protection Regulation (GDPR): A Practical
  Guide}.
\newblock Springer, 1st edition, 2017.

\bibitem{DBLP:conf/eccv/XiangDH20}
L.~Xiang, G.~Ding, and J.~Han.
\newblock Learning from multiple experts: Self-paced knowledge distillation for
  long-tailed classification.
\newblock In {\em Computer Vision - {ECCV} 2020 - 16th European Conference,
  Glasgow, UK, August 23-28, 2020, Proceedings, Part {V}}, pages 247--263,
  2020.

\bibitem{casiawebface}
D.~Yi, Z.~Lei, S.~Liao, and S.~Z. Li.
\newblock Learning face representation from scratch.
\newblock {\em CoRR}, abs/1411.7923, 2014.

\bibitem{DBLP:journals/ijcv/ZhangHZM19}
D.~Zhang, J.~Han, L.~Zhao, and D.~Meng.
\newblock Leveraging prior-knowledge for weakly supervised object detection
  under a collaborative self-paced curriculum learning framework.
\newblock {\em Int. J. Comput. Vis.}, 127(4):363--380, 2019.

\bibitem{MIPGAN}
H.~Zhang, S.~Venkatesh, R.~Ramachandra, K.~B. Raja, N.~Damer, and C.~Busch.
\newblock {MIPGAN} - generating strong and high quality morphing attacks using
  identity prior driven {GAN}.
\newblock {\em {IEEE} Trans. Biom. Behav. Identity Sci.}, 3(3):365--383, 2021.

\bibitem{mtcnn}
K.~Zhang, Z.~Zhang, Z.~Li, and Y.~Qiao.
\newblock Joint face detection and alignment using multitask cascaded
  convolutional networks.
\newblock {\em {IEEE} Signal Process. Lett.}, 23(10):1499--1503, 2016.

\bibitem{DBLP:conf/eccv/ZhangYLYYSL20}
Y.~Zhang, Z.~Yin, Y.~Li, G.~Yin, J.~Yan, J.~Shao, and Z.~Liu.
\newblock Celeba-spoof: Large-scale face anti-spoofing dataset with rich
  annotations.
\newblock In {\em Computer Vision - {ECCV} 2020 - 16th European Conference,
  Glasgow, UK, August 23-28, 2020, Proceedings, Part {XII}}, volume 12357 of
  {\em Lecture Notes in Computer Science}, pages 70--85. Springer, 2020.

\bibitem{DBLP:conf/iclr/ZongSMCLCC18}
B.~Zong, Q.~Song, M.~R. Min, W.~Cheng, C.~Lumezanu, D.~Cho, and H.~Chen.
\newblock Deep autoencoding gaussian mixture model for unsupervised anomaly
  detection.
\newblock In {\em 6th International Conference on Learning Representations,
  {ICLR} 2018, Vancouver, BC, Canada, April 30 - May 3, 2018, Conference Track
  Proceedings}, 2018.

\end{thebibliography}
}

\end{document}